\RequirePackage{snapshot}

\documentclass[shortAfour,sageh]{sagej}

\usepackage[utf8]{inputenc}

\usepackage[hidelinks]{hyperref}

\usepackage{url}
\usepackage{graphicx}
\usepackage[usenames,dvipsnames]{xcolor}
\usepackage[draft]{fixme}
\fxsetup{theme=color}
\usepackage{xspace}

\usepackage{amsmath}
\usepackage{amssymb}
\usepackage{textcomp}

\usepackage{booktabs}
\usepackage{threeparttable}
\usepackage{multirow}

\usepackage[capitalize]{cleveref}

\usepackage{tikz}
\usetikzlibrary{arrows}
\usetikzlibrary{positioning,calc}
\usetikzlibrary{decorations.pathreplacing}
\usetikzlibrary{decorations.markings}
\usetikzlibrary{fit}
\usetikzlibrary{shapes.callouts}
\usetikzlibrary{shapes.geometric}
\usetikzlibrary{matrix}
\providecommand{\acro}{CENTAURO\xspace}

\usepackage{pgfplots}
\pgfplotsset{compat=1.9}

\usepackage{epstopdf}

\usepackage{placeins}
\usepackage{dblfloatfix}

\newcommand\RGBD{\mbox{RGB-D}\xspace}

\colorlet{bg}{white}

\usepackage{pgfplotstable}
\usepackage{colortbl}

\newcommand{\findmax}[3]{
    \pgfplotstablevertcat{\datatable}{#1}
    \pgfplotstablecreatecol[
    create col/expr={    \pgfplotstablerow
    }]{rownumber}\datatable
    \pgfplotstablesort[sort key={#2},sort cmp={float >}]{\sorted}{\datatable}    \pgfplotstablegetelem{0}{rownumber}\of{\sorted}    \pgfmathtruncatemacro#3{\pgfplotsretval}
    \pgfplotstableclear{\datatable}
}

\pgfplotstableset{
    highlight/.style={
      postproc cell content/.append style={
        /pgfplots/table/@cell content={$\mathbf{\pgfmathparse{0.01 * ##1}\pgfmathprintnumber[assume math mode=true,precision=3]{\pgfmathresult}}$},
      },
    },
    typewriter/.style={
      string replace*={_}{\_},
      postproc cell content/.append style={
        /pgfplots/table/@cell content/.add={\ttfamily}{},
      }
    },
    highlight col max/.code 2 args={
        \findmax{#1}{#2}{\maxval}
        \edef\setstyles{\noexpand\pgfplotstableset{
                every row \maxval\noexpand\space column #2/.style={
                    highlight
                }
            }
        }\setstyles
    },
    zero to dash/.style={
        preproc cell content/.code={             \pgfkeysgetvalue{/pgfplots/table/@unprocessed cell content}\pgfmathresult             \pgfmathparse{(\pgfmathresult==0?int(1):int(0))}
             \ifnum\pgfmathresult>0\pgfkeyssetvalue{/pgfplots/table/@cell content}{}\fi
        },
        empty cells with={--}    }
}

\usepackage[tightpage]{preview}
\newenvironment{maybepreview}{\noindent\ignorespaces}{\par\noindent\ignorespacesafterend}

\ifPreview
\PreviewMacro[{*[][]{}}]{\includegraphics}
\else
\newcounter{includegraphicspage}
\LetLtxMacro\latexincludegraphics\includegraphics
\renewcommand{\includegraphics}[2][1]{\latexincludegraphics[page=\arabic{includegraphicspage}]{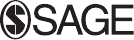}\stepcounter{includegraphicspage}}
\fi

\def\journalname{Int. Journal of Robotics Research}
\def\volumeyear{2017}
\def\volumenumber{37}
\def\issuenumber{4-5}
\def\startpage{437}
\def\endpage{451 (accepted version)}

\runninghead{Schwarz, Milan, Periyasamy, and Behnke}

\title{RGB-D Object Detection and Semantic Segmentation for Autonomous Manipulation in Clutter}

\author{Max Schwarz\affilnum{1}, Anton Milan\affilnum{2}, Arul Selvam Periyasamy\affilnum{1}, and Sven Behnke\affilnum{1}}

\affiliation{\affilnum{1}University of Bonn\\
\affilnum{2}University of Adelaide}

\corrauth{Max Schwarz, University of Bonn,
Computer Science Institute VI,
Friedrich-Ebert-Allee 144,
53113 Bonn, Germany}

\email{max.schwarz@uni-bonn.de}

\begin{document}

\begin{abstract}

Autonomous robotic manipulation in clutter is challenging.
A large variety of objects must be perceived in complex scenes, where they are partially occluded and embedded among many distractors, often in restricted spaces.
To tackle these challenges, we developed a deep-learning approach that combines object detection and semantic
segmentation.
The manipulation scenes are captured with RGB-D cameras, for which we developed a depth fusion method.
Employing pretrained features makes learning from small annotated robotic data sets possible.
We evaluate our approach on two challenging data sets: one captured for the Amazon Picking Challenge 2016, where our team NimbRo came in second in the Stowing and third in the Picking task, and one captured in disaster-response scenarios.
The experiments show that object detection and semantic segmentation complement each other and can be combined to yield reliable object perception.

\end{abstract}

\keywords{Deep learning, object perception, RGB-D camera, transfer learning, object detection, semantic segmentation}

\maketitle

\section{Introduction}
\label{sec:introduction}

Robots are increasingly deployed in unstructured and cluttered domains,
including households and disaster scenarios.
To perform complex tasks autonomously, reliable perception of the environment
is crucial. Different tasks may require different levels of 
cognition. In some cases, it may be sufficient to classify certain 
structures and objects as obstacles in order to avoid them. In others, a 
more fine-grained recognition is necessary, for instance to determine 
whether a specific object is present in the scene. For more 
sophisticated interaction, such as grasping and manipulating real-world 
objects, a more precise scene understanding including object detection 
and pixel-wise semantic segmentation is essential.

Over the past few years, research in all these domains has shown 
remarkable progress. This success is largely due to the rapid 
development of deep learning techniques that allow for end-to-end 
learning from examples, without the need for designing handcrafted 
features or introducing complex priors. 
Somewhat surprisingly, there are not many working examples to date 
that employ deep models in real-time robotic systems. 
In this paper, we first demonstrate the application 
of deep learning methods to the task of bin-picking for warehouse 
automation. This particular problem setting has unique properties: While 
the surrounding environment is usually very structured---boxes, pallets 
and shelves---the sheer number and diversity of objects that need to be 
recognized and manipulated as well as their chaotic arrangement and spatial restrictions pose daring challenges to overcome.

In addition to bin-picking, we also validate our approach in disaster-response 
scenarios. Contrary to bin-picking, this setting is much less 
structured, with highly varying and cluttered backgrounds. 
These scenes may include many unknown objects. 

\begin{figure}[t]
 \begin{maybepreview} \centering
 \includegraphics[width=\linewidth]{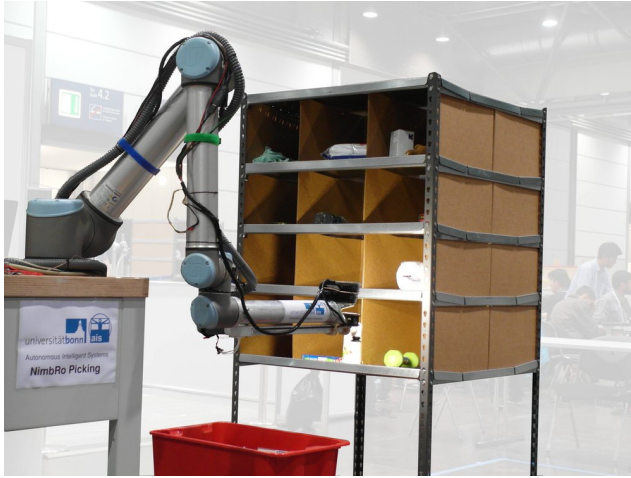}
 \end{maybepreview}
 \caption{Picking objects from the APC shelf.}
\end{figure}

Despite their remarkable success, deep learning methods exhibit at least 
one major limitation. Due to a large number of parameters, they 
typically require vast amounts of hand-labeled training data to learn the features required for solving the task.  
To overcome this limitation, we follow the  
 transfer learning approach (cf. e.g. \citet{girshick2014rich, Pinheiro:2015:CVPR, Lin:2017:CVPR}), 
 where the initial layers are pretrained on large
collections of related, already available images. These early layers are usually
responsible for extracting meaningful features like edges, corners, or
other simple structures that are often present in natural images.
Then, merely the few final layers that perform the high-level task of 
detection or segmentation need to be adjusted---or finetuned---to 
the task at hand.
This strategy enables us to exploit the power of deep neural 
networks for robotic applications with only little amounts of additional training data.

To summarize, our main contributions are as follows.

\begin{enumerate}
\item We develop two deep-learning based object perception methods
       that employ transfer learning to learn from few annotated examples
       (\cref{sec:perception}).
 \item Both deep-learning techniques are integrated into a real-life robotic system (\cref{sec:apc}).
 \item We present a simple, yet effective technique to fuse three separate sources of depth information, thereby improving the overall accuracy of depth measurements (\cref{sec:rgbd-preprocessing}, \Cref{fig:fusion}).
       The resulting depth measurements are integrated into both perception methods.
 \item Experimental results of the proposed methods are presented on two scenarios, 
 bin-picking (\cref{sec:apc}) and disaster response (\cref{sec:centauro}), showing their validity and generality.
\end{enumerate}

Finally, we discuss lessons learned (\cref{sec:lessons-learned}) from our preparation and the competition at
the Amazon Picking challenge.

\section{Related Work}
\label{sec:related-work}

\begin{figure*}[t]
 \begin{maybepreview}
 \centering
 \includegraphics[width=.9\linewidth]{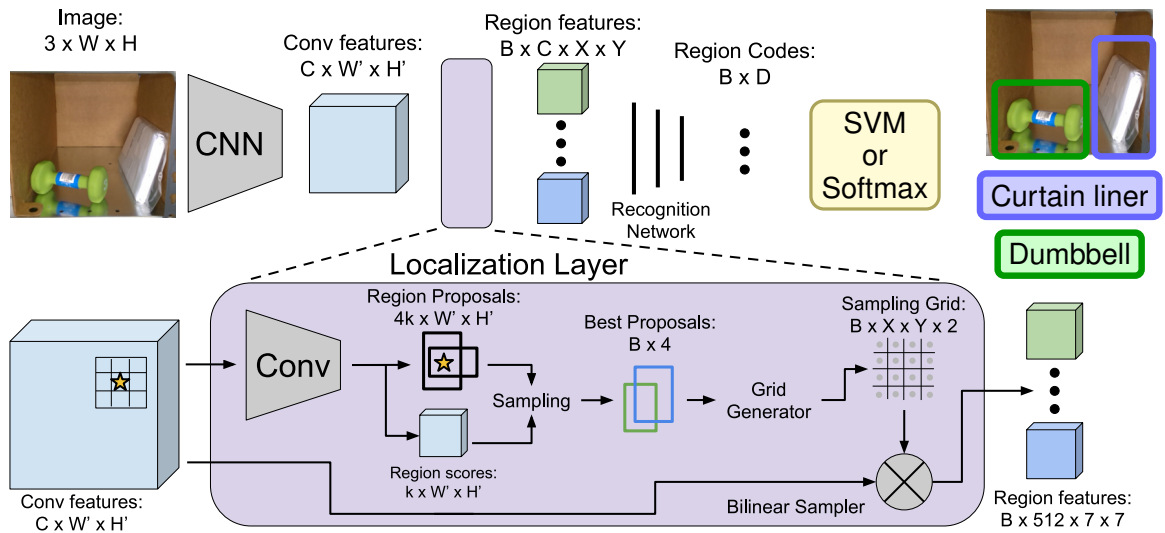}
 \end{maybepreview}
 \caption{Architecture of the RGB object detection pipeline. Adapted from~\cite{Johnson:2016:CVPR}.
 The input size WxH is flexible, for the APC situation we used 720x405.
 VGG-16 produces C=512 feature maps after the last convolution.
 B=1000 bounding boxes are extracted using non-maximum suppression.
 }
 \label{fig:densecap-architecture}
\end{figure*}

Even though early ideas of deep neural networks date back several 
decades~\citep{Ivakhnenko:1966, LeCun:1989:CM, behnke2003hierarchical}, it was not until recently
that they gained immense popularity in numerous applications including 
machine translation~\citep{Sutskever:2014:NIPS}, speech
recognition~\citep{Graves:2013:ICASSP}, and computer
vision~\citep{Krizhevsky:2012:NIPS}. Powered by the
parallel computing architectures present in modern GPUs as well as the 
availability of labeled data~\citep{Russakovsky:2015:IJCV, Krishna:2016:visual, Song:2015:CVPR},
deep learning methods now easily outperform 
traditional approaches in tasks like image 
classification~\citep{Krizhevsky:2012:NIPS}, object
detection~\citep{Johnson:2016:CVPR, Liu:2016:ECCV}, or semantic
segmentation~\citep{Chen:2014:ICLR, Lin:2017:CVPR}. The core of such
methods typically constitutes a multi-layer architecture, where each 
layer consists of a series of convolutional operations, passed through a non-linear 
activation function, such as e.g. rectified linear unit (ReLU), followed 
by spatial maximum pooling~\citep{scherer2010evaluation} to produce local deformation invariance. Perhaps 
one of the most popular of such networks is the architecture proposed 
by \citet{Krizhevsky:2012:NIPS}, often referred to as AlexNet. Although
initially employed for image classification, it has been since adapted 
to various other tasks including image 
captioning~\citep{Karpathy:2015:CVPR}, pedestrian
detection~\citep{Girshick:2015:ICCV}, and semantic
segmentation~\citep{Long:2015:CVPR}. The original AlexNet consists of five
convolutional and three fully-connected layers. Other prominent deep 
networks include the 16-layer VGG~\citep{Simonyan:2014:VGG}, Inception
with 22 layers~\citep{Szegedy:2014:CoRR}, and more recently the so-called
Deep Residual Networks with skip connections and up to 152 
layers~\citep{He:2016:ResNet}.
Other works in the area of semantic segmentation include \cite{badrinarayanan2015segnet},
who use a multi-stage encoder-decoder architecture that first uses maximum
pooling to reduce spatial resolution and later upsample the segmentation results
using the indices of the local pooling maxima.
For object detection, one line of work considers the task as region proposal
followed by classification and scoring. \citet{girshick2014rich} process
external region proposals using RoI pooling to reshape intermediate CNN feature
maps to a fixed size. To increase performance, all regions may be processed
in a single forward pass \citep{Girshick:2015:ICCV}. Finally, region proposal
networks that regress from anchors to regions of interest are integrated into
the Faster R-CNN detection framework~\citep{ren2015faster}.

Our approach is
mainly based on two methods: the OverFeat feature extractor introduced 
by \citet{Sermanet:2014:ICLR}, as well as the DenseCap region detection
and captioning approach of \citet{Johnson:2016:CVPR}. In particular,
we adopt both approaches to detect and segment objects for robotic
perception and manipulation in new settings, namely bin picking and disaster-response tasks.
To that end, we rely on transfer learning, a method for adapting pre-trained models for a specific task. 
Note that this training approach is not entirely new and in fact has been followed in several of the above works.
For example, R-CNN~\citep{girshick2014rich} starts from ImageNet features and fine-tunes a CNN for object detection.
\Citet{Pinheiro:2015:CVPR} exploit large amounts of image-labeled training data to train a CNN for a more fine-grained task of semantic segmentation.
\Citet{schwarz2015rgb} used pretrained features and depth preprocessing to recognize RGB-D objects
and to estimate their pose.
In this work, we apply transfer learning to robotic perception and manipulation scenarios.
Recently, combinations of object detection and semantic segmentation have been
introduced, such as Mask R-CNN \citep{he2017mask}, which predicts local 
segmentation masks for each detected object. In contrast, our work
combines results after separate object detection and semantic segmentation,
allowing independent training. Nevertheless, combining both approaches in a
single network is both elegant and effective.

Bin picking is one of the classical problems in robotics and has been 
investigated by many research groups in the last three decades, 
e.g.~\citep{bucholz_2014, nieuwenhuisen2013mobile, pretto2013flexible,
domae_2014, drost_2010, Berner:ICIP2013, martinez2015automated, holz2015real,
kaipa2016addressing, harada2016iterative}. In these works, often 
simplifying conditions are exploited, e.g. known parts of one type being 
in the bin, parts with holes that are easy to grasp by sticking fingers 
inside, flat parts, parts composed of geometric primitives, well 
textured parts, or ferrous parts that can be grasped with a magnetic 
gripper.

During the Amazon Picking Challenge (APC) 2015, 
various approaches to a more general shelf-picking 
problem have been proposed and evaluated. \Citet{correll2016lessons} 
aggregate lessons learned during the APC 2015 and present a general 
overview and statistics of the approaches. For example, 36\,\% of all 
teams (seven of the top ten teams) used suction for manipulating the 
objects.

\Citet{APC_RSS_2016} describe their winning system for APC 2015. 
Mechanically, the robot consists of a mobile base and a 7-DOF arm to 
reach all shelf bins comfortably. In contrast, our system uses a larger 
arm and can thus operate without a mobile base (see \cref{sec:design}). 
The endeffector of \citet{APC_RSS_2016} is designed as a fixed suction
gripper, which can execute top and side picks. Front picks are, however,
not possible. The object perception system is described in detail by
\citet{Jonschkowski:2016:IROS}. A single RGB-D camera captures the 
scene. Six hand-crafted features are extracted for each pixel, including 
color and geometry-based features. The features are then used in a 
histogram backprojection scheme to estimate the posterior probability 
for a particular object class. The target segment is found by searching 
for the pixel with the maximum probability. After fitting a 3D bounding 
box, top or side grasps are selected heuristically.
The team performed very well at APC 
2015 and reached 148 out of 190 points.

\Citet{yu2016summary} reached second place with Team MIT in the APC 
2015. Their system uses a stationary industrial arm and a hybrid 
suction/gripping endeffector. The industrial arm provides high accuracy 
and also high speed. Similar to our approach, an Intel RealSense sensor 
mounted on the wrist is used for capturing views of the bin scenes 
(together with two base-mounted Kinect2 sensors). A depth-only GPU-based 
instance registration approach is used to determine object poses.
Team MIT achieved 88 points in the competition.

In contrast to the first edition, the 2016 Amazon Picking Challenge 
introduced more difficult objects (e.g. the heavy 3\,lb dumbbell), 
increased the difficulty in the arrangements and the number of items per 
bin, and introduced the new stowing task.
\Citet{Leitner:2017:ICRA} introduce a physical picking benchmark to ease 
reproducibility in warehouse automation research and 
present their own approach as a baseline.

\Citet{Hernandez:2016:APC} reached first place in both the picking and the
stowing task in the APC 2016. Their system consists of a large industrial
7-DOF arm mounted on a horizontal rail, resulting in eight degrees of freedom
in the arm and base.
The gripper is a complex custom design, allowing both suction and pinch grasps.
Like our design described in \cref{sec:design}, the suction cup can
be bent to facilitate top, side, and frontal grasps.
Object perception is based on \RGBD measurements from an Ensenso 3D camera.
The authors report problems with reflections and noise, and therefore built in
heuristics to reject false registrations.
In contrast, we added a second \RGBD camera
to be able to filter out false measurements (see \cref{sec:rgbd-preprocessing}).
Similarly to our architecture, object detection is carried out using an approach
based on Faster R-CNN~\citep{ren2015faster}. After detection, object poses
are estimated using a point cloud registration method.
For grasp planning, primitive shapes are fitted to find grasp candidate spots.
Candidates are then filtered using reachability measures.
For deformable objects, a simple measurement-based heuristic is used, similar to our
approach.

\cite{Zeng:2017:ICRA} describe the approach of Team MIT-Princeton, who performed successfully in the APC 2016 reaching 3rd and 4th place. 
Like \Citet{Hernandez:2016:APC}, a high-precision industrial robot arm with a combination of a two-finger gripper and a suction cup was used for object manipulation.
The perception pipeline included segmentation and 6D pose estimation of objects based on a multi-view point cloud.
Segmentation is addressed using a VGG-type CNN~\citep{Simonyan:2014:VGG} in each of the 15-18 RGB-D frames, which are then combined to produce a semantic 3D point cloud.
Pose estimation is then performed on the dense point cloud by aligning 3D-scanned objects using a modified ICP algorithm.
Note that over 130,000 training images were used for their system, which is about three orders of magnitude more than in our approach.

\section{Methods}
\label{sec:perception}
\begin{figure*}[t]
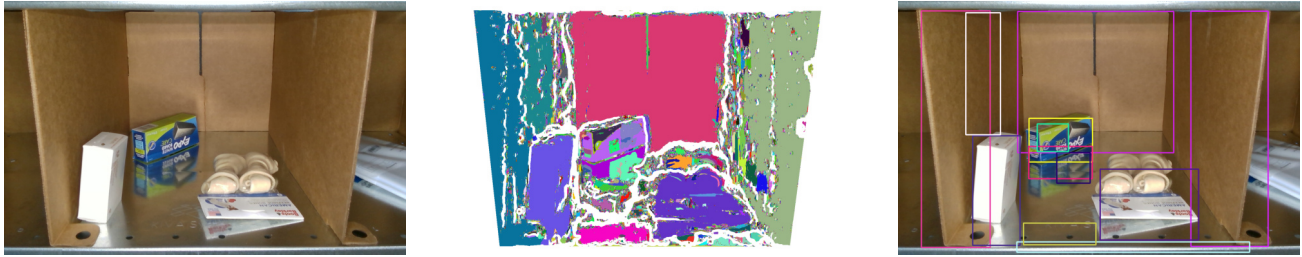

 \begin{maybepreview} \centering
 \includegraphics[width=.31\linewidth]{images/perception_examples/frame3/rgb-cropped}
 \hfill
 \includegraphics[width=.31\linewidth]{images/perception_examples/frame3/proposal_vis-cropped}
  \hfill
 \includegraphics[width=.31\linewidth]{images/perception_examples/frame3/proposals-cropped}
 \end{maybepreview}
 \caption{RGB-D based additional region proposals. Left: RGB frame.
 Center: Regions labeled using the connected components algorithm.
 Right: Extracted bounding box proposals.}
 \label{fig:densecap:proposals}
\end{figure*}
For perceiving objects in the vicinity of a robot, we developed two 
independent methods. The first one solves the object detection problem, 
i.e. outputs bounding boxes and object classes for each detection. The 
second one performs semantic segmentation, which provides a pixel-wise 
object category labeling.

Since training data and time are limited, it is crucial not to train from 
scratch. Instead, both methods leverage convolutional neural networks 
(CNNs) pre-trained on large image classification datasets and merely 
adapt the networks to work in the specific domain.

\subsection{Object Detection}
\label{sec:object-detection}

We extend an object detection approach based on the DenseCap 
network~\citep{Johnson:2016:CVPR}. DenseCap approaches the problem of 
dense captioning, i.e. providing detailed textual descriptions of 
interesting regions (bounding boxes) in the input image. 
\Cref{fig:densecap-architecture} shows the general architecture of the 
DenseCap network. 
In a nutshell, the network first extracts CNN features and samples 
a fixed number of proposals (1000 in our case) using an objectness
score within a region proposal network. Since the sampled regions
can be of arbitrary size and shape, the intermediate CNN feature maps are 
interpolated to fit a fixed size for each proposal. The proposals are then 
classified using a recognition network. In contrast to other recent works on object
detection \citep[e.g. Faster R-CNN,][]{ren2015faster}, the DenseCap architecture
can be trained end-to-end without the need for approximations, since its
bilinear interpolation layer is fully differentiable.
The underlying VGG-16 CNN \citep{Simonyan:2014:VGG} used for feature extraction
was pretrained on the ImageNet \citep{Russakovsky:2015:IJCV} dataset.
Afterwards, the entire pipeline was trained end-to-end on the Visual Genome 
dataset \citep{Krishna:2016:visual}.

Obviously, textual descriptions of image regions are not relevant for bin-picking 
scenarios.  However, these captions are generated from an intermediate feature 
vector representation, which is highly descriptive and therefore should
not be simply ignored.
To exploit the power of this feature representation, we use the network 
up until the captioning module for feature extraction, and replace the
language processing part by a classification component that uses a linear 
support vector machine (SVM).
As an alternative, we investigate a soft-max classifier layer, 
which allows us to fine-tune the network during training.

\begin{figure*}
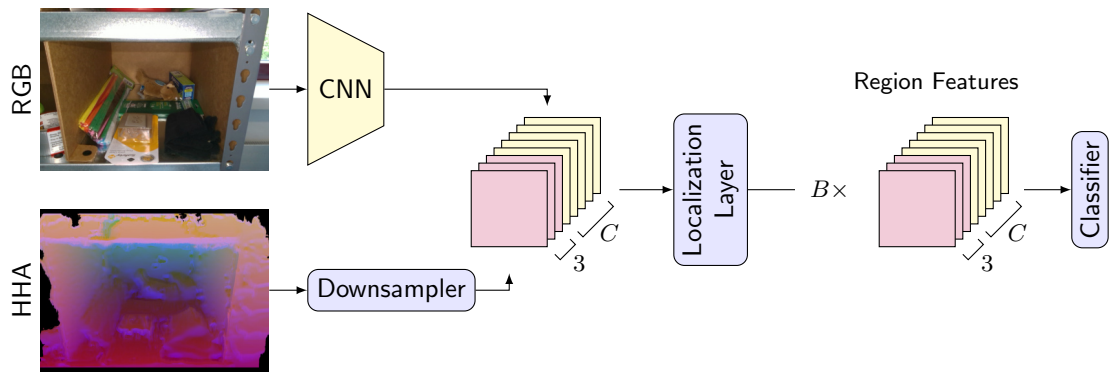

 \begin{maybepreview} \centering
 \begin{tikzpicture}[
font=\sffamily,
cnn/.style={draw=black,
minimum height=1cm, minimum width=2cm, trapezium, trapezium stretches,trapezium angle=75,shape border rotate=270
},
rgb/.style={fill=yellow!20},
depth/.style={fill=purple!20},
ncbar angle/.initial=90,
ncbar label/.initial={},
    ncbar/.style={
        to path=(\tikztostart)
        -- ($(\tikztostart)!#1!\pgfkeysvalueof{/tikz/ncbar angle}:(\tikztotarget)$)
        -- node[midway] (midwaylabel) {} ($(\tikztotarget)!($(\tikztostart)!#1!\pgfkeysvalueof{/tikz/ncbar angle}:(\tikztotarget)$)!\pgfkeysvalueof{/tikz/ncbar angle}:(\tikztostart)$)
        -- (\tikztotarget)
    },
    ncbar/.default=0.5cm,
layer/.style={fill=blue!10,draw=black,rectangle,rounded corners,align=center},
]

\node[inner sep=0] (rgb) at (0,0) {\includegraphics[width=3cm,clip,trim=200 0 200 0 0]{images/perception_examples/frame0/rgb}};
\node[inner sep=0,below=0.5cm of rgb] (hha) {\includegraphics[width=3cm,clip,trim=200 0 200 0 0]{images/perception_examples/frame0/feature_hha}};

\node[left=0 of rgb,rotate=90,anchor=south] {RGB};
\node[left=0 of hha,rotate=90,anchor=south] {HHA};

\node[cnn,rgb,right=0.5cm of rgb] (rgbcnn) {CNN};
\node[layer,right=0.5cm of hha] (hhacnn) {Downsampler};

\coordinate (merged) at ($(rgbcnn)!0.5!(hhacnn)$);

\node[align=center,draw=black,layer,rectangle, rounded corners,right=4.5cm of merged,anchor=center,rotate=90] (loclayer) {Localization\\Layer};

\node[align=center,draw=black,layer,rectangle,rounded corners,right=4.5cm of loclayer.south,anchor=center,rotate=90] (classifier) {Classifier};

\draw[-latex] (rgb) edge (rgbcnn);
\draw[-latex] (hha) edge (hhacnn);

\coordinate (featuremaps) at ($(merged)+(2.1,0)$);

\begin{scope}[shift={(featuremaps)}]
\foreach \i in {-4,...,0} {
 \draw[rgb] ($(-0.45,-0.45)-0.1*(\i,\i)$) rectangle ++(1,1);
};
\draw[black,shift={(0.1,-0.1)}] (0.95,-0.05) to [ncbar=0.1cm] (0.55,-0.45);
\node[font=\small\sffamily,shift={(0.15,-0.15)}] at (midwaylabel) {$C$};

\foreach \i in {0,...,2} {
 \draw[depth] ($(-0.55,-0.55)-0.1*(\i,\i)$) rectangle ++(1,1);
};
\draw[black,shift={(0.1,-0.1)}] (0.45,-0.55) to [ncbar=0.1cm] (0.25,-0.75);
\node[font=\small\sffamily,shift={(0.15,-0.15)}] at (midwaylabel) {$3$};

\coordinate (rgbinput) at (0.25,1.1);
\draw[-latex] (rgbcnn) -| (rgbinput);

\coordinate (hhainput) at  (-0.25,-1.1);
\draw[-latex] (hhacnn) -| (hhainput);

\end{scope}

\draw[-latex] (featuremaps) +(1.2,0) -- (loclayer);

\draw[-latex] (loclayer) edge node[midway,minimum width=3.0cm,minimum height=2cm,fill=bg] (region_featuremaps) {} (classifier);

\begin{scope}[shift={(region_featuremaps)}]
\node[font=\sffamily\small] at (-1.05,0) {$B\times$};
\begin{scope}[shift={(0.35,0)}]
	\foreach \i in {-4,...,0} {
	 \draw[rgb] ($(-0.45,-0.45)-0.1*(\i,\i)$) rectangle ++(1,1);
	};
	\draw[black,shift={(0.1,-0.1)}] (0.95,-0.05) to [ncbar=0.1cm] (0.55,-0.45);
	\node[font=\small\sffamily,shift={(0.15,-0.15)}] at (midwaylabel) {$C$};

	\foreach \i in {0,...,2} {
 		\draw[depth] ($(-0.55,-0.55)-0.1*(\i,\i)$) rectangle ++(1,1);
	};
	\draw[black,shift={(0.1,-0.1)}] (0.45,-0.55) to [ncbar=0.1cm] (0.25,-0.75);
	\node[font=\small\sffamily,shift={(0.15,-0.15)}] at (midwaylabel) {$3$};
    \node[font=\small\sffamily] at (0,1.4) {Region Features};
\end{scope}

\end{scope}

\end{tikzpicture}  \end{maybepreview}
 \caption[Downsampled HHA as additional feature maps]{Detection pipeline with CNN features from RGB and downsampled HHA-encoded depth.
 $C$ denotes the number of CNN feature maps after the last convolutional layer (512 for VGG-16). The internal proposal generator produces $B$ proposals (1000).
 }
 \label{fig:method:depth:downsample}
\end{figure*}

\subsubsection{Linear SVM.}
\label{sec:linear-svm}
In the first case, we remove the language generation model and replace 
it with a linear SVM for classification. We also introduce two primitive 
features based on depth: The predicted bounding box is projected into 3D 
using the depth value of its center. The metric area and size are then 
concatenated to the CNN features. Since linear SVMs can be trained very 
efficiently, the training can happen just-in-time before actual 
perception, exploiting the fact that the set of possible objects in the 
bin is known. Note that restricting the set of classes to the ones present
in the current scene also has the side-effect that training time and 
memory usage are constant with respect to the set 
of all objects present in a warehouse. This allows us to potentially scale 
this approach to an arbitrarily large number of object categories.

The SVM is used to classify each predicted bounding box. To identify a 
single output, the bounding box with the maximum SVM response is 
selected. This ignores duplicate objects, but since the goal is to 
retrieve only one object at a time, this reduction is permissible.

\subsubsection{Softmax Classification.}
\label{sec:finetuning}
For finetuning the network, we use a soft-max classification layer 
instead of the SVM. All layers except the initial CNN layers (see 
\cref{fig:densecap-architecture}) are optimized for the task at hand. 
Contrary to SVM classification, the softmax layer predicts confidences
over all object classes. In the bin-picking scenario, the bounding box
with the highest score in the desired object class is produced as the
final output, i.e. the object to pick.

\subsubsection{Incorporating Depth.}
\label{sec:densecap:depth}

The existing object detection network does not make use of depth measurements.
Here, we investigate several methods for incorporating depth into the network.

As with all architectures based on R-CNN, it is straightforward to classify
bounding boxes generated from an external proposal generator. One way to include
depth information is therefore to use an external RGB-D proposal generator.
To this end, we augment the 
network-generated proposals with proposals from a connected components 
algorithm running on the RGB and depth frames (see 
\cref{fig:densecap:proposals}). Two pixels are deemed connected if they 
do not differ more than a threshold in terms of 3D position (5\,mm), normal
angle ($50^\circ$), saturation, and color ($10$). Final bounding boxes are extracted from
regions with an area above a predefined threshold (10,000 pixels for 1920$\times$1080 input).

\begin{figure*}
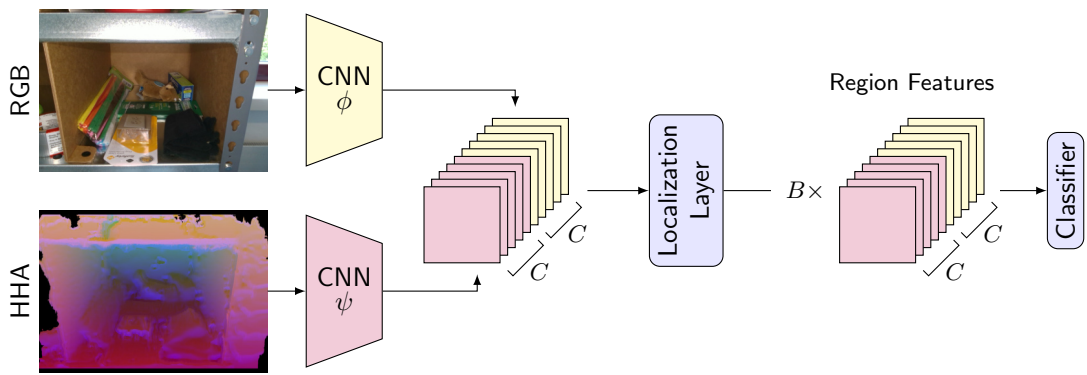

 \begin{maybepreview} \centering
 \begin{tikzpicture}[
font=\sffamily,
cnn/.style={draw=black,
minimum height=1cm, minimum width=2cm, trapezium, trapezium stretches,trapezium angle=75,shape border rotate=270
},
rgb/.style={fill=yellow!20},
depth/.style={fill=purple!20},
ncbar angle/.initial=90,
ncbar label/.initial={},
    ncbar/.style={
        to path=(\tikztostart)
        -- ($(\tikztostart)!#1!\pgfkeysvalueof{/tikz/ncbar angle}:(\tikztotarget)$)
        -- node[midway] (midwaylabel) {} ($(\tikztotarget)!($(\tikztostart)!#1!\pgfkeysvalueof{/tikz/ncbar angle}:(\tikztotarget)$)!\pgfkeysvalueof{/tikz/ncbar angle}:(\tikztostart)$)
        -- (\tikztotarget)
    },
    ncbar/.default=0.5cm,
layer/.style={fill=blue!10,draw=black,rectangle,rounded corners,align=center},
]

\node[inner sep=0] (rgb) at (0,0) {\includegraphics[width=3cm,clip,trim=200 0 200 0 0]{images/perception_examples/frame0/rgb}};
\node[inner sep=0,below=0.5cm of rgb] (hha) {\includegraphics[width=3cm,clip,trim=200 0 200 0 0]{images/perception_examples/frame0/feature_hha}};
\node[left=0 of rgb,rotate=90,anchor=south] {RGB};
\node[left=0 of hha,rotate=90,anchor=south] {HHA};

\node[cnn,rgb,right=0.5cm of rgb,align=center] (rgbcnn) {CNN\\[-0.1cm]$\phi$};
\node[cnn,depth,right=0.5cm of hha,align=center] (hhacnn) {CNN\\[-0.1cm]$\psi$};

\coordinate (merged) at ($(rgbcnn)!0.5!(hhacnn)$);

\node[align=center,draw=black,layer,rectangle, rounded corners,right=4.5cm of merged,anchor=center,rotate=90] (loclayer) {Localization\\Layer};

\node[align=center,draw=black,layer,rectangle,rounded corners,right=4.5cm of loclayer.south,anchor=center,rotate=90] (classifier) {Classifier};

\draw[-latex] (rgb) edge (rgbcnn);
\draw[-latex] (hha) edge (hhacnn);

\coordinate (featuremaps) at ($(merged)+(2,0)$);

\begin{scope}[shift={(featuremaps)}]
\foreach \i in {-4,...,0} {
 \draw[rgb] ($(-0.45,-0.45)-0.1*(\i,\i)$) rectangle ++(1,1);
};
\draw[black,shift={(0.1,-0.1)}] (0.95,-0.05) to [ncbar=0.1cm] (0.55,-0.45);
\node[font=\small\sffamily,shift={(0.15,-0.15)}] at (midwaylabel) {$C$};

\foreach \i in {0,...,4} {
 \draw[depth] ($(-0.55,-0.55)-0.1*(\i,\i)$) rectangle ++(1,1);
};
\draw[black,shift={(0.1,-0.1)}] (0.45,-0.55) to [ncbar=0.1cm] (0.05,-0.95);
\node[font=\small\sffamily,shift={(0.15,-0.15)}] at (midwaylabel) {$C$};

\coordinate (rgbinput) at (0.25,1.1);
\draw[-latex] (rgbcnn) -| (rgbinput);

\coordinate (hhainput) at  (-0.25,-1.1);
\draw[-latex] (hhacnn) -| (hhainput);

\end{scope}

\draw[-latex] (featuremaps) +(1.2,0) -- (loclayer);

\draw[-latex] (loclayer) edge node[midway,minimum width=3.0cm,minimum height=2cm,fill=bg] (region_featuremaps) {} (classifier);

\begin{scope}[shift={(region_featuremaps)}]
\node[font=\sffamily\small] at (-1.05,0) {$B\times$};
\begin{scope}[shift={(0.35,0)}]
	\foreach \i in {-4,...,0} {
	 \draw[rgb] ($(-0.45,-0.45)-0.1*(\i,\i)$) rectangle ++(1,1);
	};
	\draw[black,shift={(0.1,-0.1)}] (0.95,-0.05) to [ncbar=0.1cm] (0.55,-0.45);
	\node[font=\small\sffamily,shift={(0.15,-0.15)}] at (midwaylabel) {$C$};

	\foreach \i in {0,...,4} {
	 \draw[depth] ($(-0.55,-0.55)-0.1*(\i,\i)$) rectangle ++(1,1);
	};
	\draw[black,shift={(0.1,-0.1)}] (0.45,-0.55) to [ncbar=0.1cm] (0.05,-0.95);
	\node[font=\small\sffamily,shift={(0.15,-0.15)}] at (midwaylabel) {$C$};
    \node[font=\small\sffamily] at (0,1.4) {Region Features};
\end{scope}

\end{scope}

\end{tikzpicture}  \end{maybepreview}
 \caption[Concatenated RGB-D CNN features]{Detection pipeline with concatenated CNN features from RGB and HHA-encoded depth.
 $C$ denotes the number of CNN feature maps after the last convolutional layer (512 for VGG-16). The internal proposal generator produces $B$ proposals (1000).
 For the Cross Modal Distillation approach, CNN $\psi$ is trained to imitate the pretrained CNN $\phi$.
 }
 \label{fig:method:depth:2cnn}
\end{figure*}

A second possibility is to treat depth as an additional mid-level feature.
For this purpose, we use the popular three-channel HHA
encoding~\citep{Gupta:2014:ECCV}, which augments depth with two geometric
features (height above ground and angle to gravity).
We downsample the HHA map and concatenate it to the feature maps generated by
the pretrained first convolutional layers
(see \cref{fig:method:depth:downsample}). Furthermore, we can also use the same
pretrained CNN to extract higher-level features from HHA, as shown in
\cref{fig:method:depth:2cnn}.

Finally, \citet{gupta2016cross} propose to use an RGB reference network to
generate the training data needed for the other modality, a technique they call
Cross Modal Distillation. In essence, the pretrained RGB network $\phi$ computes
a feed-forward pass on the RGB frame $I_s$, generating the target feature maps $\phi(I_s)$. A back-propagation step
then trains the depth network $\psi$ to imitate these features on the corresponding depth frame $I_d$, minimizing the objective
\begin{equation}
 \min_{W_d} \sum_{(I_s, I_d) \in U_{s,d}} || \psi(I_d) - \phi(I_s) ||^2,
\end{equation}
where $W_d$ are the weights of the depth network, and $U_{s,d}$ is the training
set of RGB and depth frame pairs.
Note that no annotation is necessary on $U_{s,d}$,
so any RGB-D video (ideally of the target domain) can be used to perform the
supervision transfer. In our case, the (annotated) training set
is used for distillation, since additional unlabeled RGB-D sequences of the
target domain are not available.

After the initial Cross Modal Distillation training, the trained network can
be used in place of the RGB network for depth feature extraction
(see \cref{fig:method:depth:2cnn}).

\subsubsection{Implementation Details.}
As in the original DenseCap work, the ADAM optimizer \citep{kingma2014adam}
is used for training the network with parameters $\beta_1=0.9$, $\beta_2=0.999$
and $\epsilon=10^{-8}$.
However, we adapt a custom learning rate
schedule to dampen training oscillations at the end of training (annealing):
The learning rate starts at $1\cdot10^{-5}$ and is kept constant for 15 epochs, then
linearly lowered to $1\cdot10^{-6}$ during the next 85 epochs. At 200 epochs, the rate
is lowered to $5\cdot10^{-7}$, and finally to $1\cdot10^{-7}$ at 250 epochs.
To prevent overfitting, 50\% dropout is used throughout the entire network.
As in the original DenseCap pipeline, input images are scaled such that the
longest side has 720 pixels.

\begin{figure*}[!t]
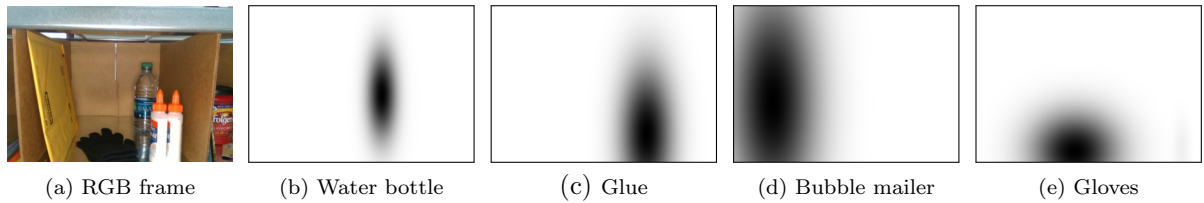

 \begin{maybepreview} \centering
 \setlength{\tabcolsep}{3pt}
 \setlength{\fboxsep}{0pt}
 \begin{tabular}{ccccc}
  \includegraphics[height=.12\linewidth,clip,trim=200 100 200 0]{images/densecap/frame1/rgb}  &
  \fbox{\includegraphics[height=.12\linewidth,clip,trim=200 100 200 0]{images/densecap/frame1/inverted/prob_dasani_water_bottle}}  &
  \fbox{\includegraphics[height=.12\linewidth,clip,trim=200 100 200 0]{images/densecap/frame1/inverted/prob_elmers_washable_no_run_school_glue}} &
  \fbox{\includegraphics[height=.12\linewidth,clip,trim=200 100 200 0]{images/densecap/frame1/inverted/prob_scotch_bubble_mailer}} &
  \fbox{\includegraphics[height=.12\linewidth,clip,trim=200 100 200 0]{images/densecap/frame1/inverted/prob_womens_knit_gloves}} \\
  \footnotesize(a) RGB frame & \footnotesize(b) Water bottle & (c) \footnotesize Glue & \footnotesize (d) Bubble mailer & \footnotesize (e) Gloves \\
 \end{tabular}
 \end{maybepreview}
 \caption{DenseCap probability estimates for an example frame. The objectness in each pixel is approximated by rendering a Gaussian that corresponds to the bounding box center and extent.}
 \label{fig:dc_gauss}
\end{figure*}

\subsection{Semantic Segmentation}
\label{sec:semantic-segmentation}

Manipulation of real-world objects requires a precise object localization.
Therefore, we also 
investigated pixel-level segmentation approaches in the context of
robot manipulation tasks.
As opposed to object detection, which only provides a rather coarse
estimate of object location in terms of a bounding box, segmentation
offers a much more detailed representation by classifying each pixel as
one of the known categories.

\begin{figure}
  \begin{maybepreview}  \includegraphics[width=\linewidth]{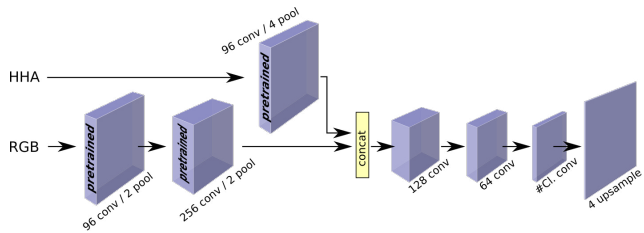}
  \end{maybepreview}
  \caption{Our network architecture for semantic object segmentation. }
  \label{fig:segmentation-cnn}
\end{figure}

Our segmentation network is based on the work of
\citet{Husain:2016:RAL} and consists of six convolution layers, as shown in 
Figure~\ref{fig:segmentation-cnn}. Only the last three layers of the network are trained 
on the domain specific dataset.
For RGB data, the first two convolutional layers from the OverFeat
network~\citep{Sermanet:2014:ICLR} are applied.
OverFeat was
trained on the ImageNet dataset~\citep{Russakovsky:2015:IJCV} 
consisting of over one million images, each of which was labeled with 
one of 1000 available categories.
For depth data, we also use the HHA encoding, similar to the object detection
pipeline. Since the OverFeat network was trained in the RGB domain, and its higher-level
features are not meaningful on HHA, we only use the first convolutional layer
of the network for HHA. Still, CNN pretraining on RGB can be used on HHA, as
demonstrated by \citet{Gupta:2014:ECCV}.
The extracted RGB and HHA features are then concatenated.
The dimension of the final convolution layer is adjusted for each task to reflect the
number of classes (\#Cl.) present in the dataset.

\subsubsection{Implementation Details.}
The network is finetuned using stochastic gradient descent.
We follow~\citet{Husain:2016:RAL} and set the
learning rate to $10^{-3}$, use a momentum of $0.9$, and a $50$\%
 dropout in the final three layers. The learning rate is set to decay 
 over time at a rate of $10^{-4}$.
 
Both detection and segmentation methods were implemented using the Torch7 deep learning
framework\footnote{\url{http://torch.ch}} and integrated
into a real-time robotics system based on ROS~\citep{quigley2009ros}.
We hope that our implementation, which is publicly available\footnote{\url{http://ais.uni-bonn.de/apc2016/}}, will lower the burden for
other researchers to apply deep learning methods for robotic tasks.

\subsection{Combining Detection and Segmentation}
\label{sec:detection-segmentation-combination}

During the APC competition, we used a combination of the SVM object 
detection approach 
and the semantic segmentation. 
In particular, the bounding boxes predicted by the 
object detection network were rendered with a logistic estimate of their 
probability and averaged over all classes. 
This process produced a ``probability map'' 
that behaved similar to a pixel-wise posterior. In the next step, the detection
probability map was multiplied element-wise with the class probabilities 
determined in semantic segmentation.
A per-pixel max-probability decision then resulted in the final 
segmentation mask used in the rest of the pipeline.

After the APC, we replaced the hard bounding box rendering with a soft 
Gaussian whose mean and covariance were derived from the box location
and size, respectively, cf. \cref{fig:dc_gauss} for an illustration.
This yielded better results, because such a representation
matches the actual object shape more closely than an axis-aligned
bounding box. The Gaussian blobs for all detections are accumulated and the resulting
map $P_{\textrm{det}}$ is normalized, i.e. scaled so that the maximum equals to one. To allow for
detection mistakes, we introduce a weak prior that accounts for false negatives. 
The final combined posterior is computed as
\begin{equation}
 P_{\textrm{combined}} = P_{\textrm{seg}} (0.1 + 0.9 P_{\textrm{det}}),
\end{equation}
where $P_{\textrm{seg}}$ is the posterior resulting from
semantic segmentation and $P_{\textrm{det}}$ is the estimated posterior
from object detection.

While this combination is relatively straightforward (note that the product assumes
conditional independence) and the shape approximations by Gaussian masks are rather coarse, this strategy yields a consistent
increase in performance nonetheless (see \cref{sec:eval:object_detection}).

\section{Application to Bin-Picking}
\label{sec:apc}

In July 2016, in conjunction with RoboCup, Amazon held the second 
Amazon Picking Challenge 
(APC)\footnote{ \url{http://amazonpickingchallenge.org/}}, which 
provided a platform for comparing state-of-the-art solutions and new 
developments in bin picking and stowing applications. The challenge 
consisted of two separate tasks, where contestants were required to pick 
twelve specified items out of chaotically arranged shelf boxes---and to 
stow twelve items from an unordered pile in a tote into the shelf. 
Amazon provided a set of objects from 39 categories, representing a 
large variety of challenging properties, including transparency (e.g. 
water bottle), shiny surfaces (e.g. metal or shrink wrap), deformable 
materials (e.g. textiles), black and white textureless surfaces which 
hamper reliable depth measurements, heavy objects, and mesh-like objects 
with holes that could not be grasped by using suction alone. Moreover, the 
shiny metal floors of the shelf boxes posed a considerable challenge to 
the perception systems, as all objects are also visible through their 
mirrored image. Before the run, the system was supplied with a task 
file that specified which objects should be picked as well as with all object 
locations (which shelf box they are stored in). After the run, the system was 
expected to output the new locations of the items.
For completeness, before presenting our results 
we will briefly describe our mechatronic design as 
well as motion generation and grasp selection methods that were 
developed for the APC competition.

\subsection{Robotic System}
\label{sec:design}

Our robot consists of a 6-DOF arm, a 2-DOF endeffector, a camera module, 
and a suction system.
To limit system complexity, we chose to use a stationary manipulator. 
This means the manipulation workspace has to cover the entire shelf, 
which places constraints on the possible robotic arm solutions. In our 
case, we chose the UR10 arm from Universal Robotics, because it covers 
the workspace sufficiently, is cost-effective, lightweight, and offers safety 
features such as an automatic (and reversible) stop upon contact with 
the environment.

\begin{figure}
 \begin{maybepreview} \centering
 
\begin{tikzpicture}[font=\sffamily\footnotesize]

\node[anchor=south west,inner sep=0] (image) at (0,0) {\includegraphics[width=\linewidth]{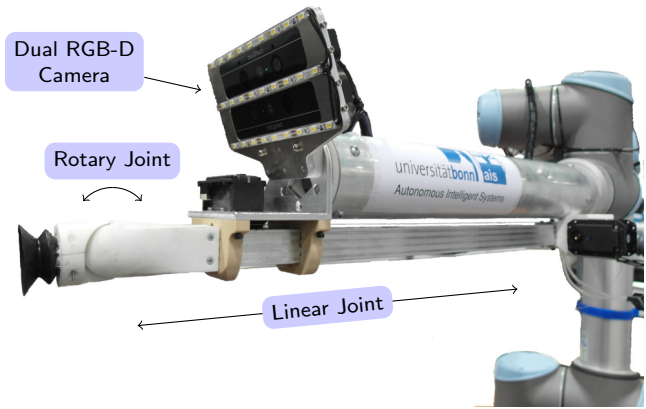}};
    \begin{scope}[x={(image.south east)},y={(image.north west)}, every node/.style={fill=blue!20, rounded corners}]

        \draw[<->] (0.2,0.2) --  (0.8,0.29) node[midway,sloped] {Linear Joint};

        \coordinate (rot) at (0.16,0.54);

        \draw[->] (rot) arc [start angle=90, end angle=40,radius=0.5cm];
        \draw[->] (rot) arc [start angle=90, end angle=140,radius=0.5cm];
        \node[above=0.1cm of rot] {Rotary Joint};

        \node[align=center] (cam) at (0.1,0.85) {Dual RGB-D\\Camera};
        \draw[->] (cam) -- (0.3,0.78);
    \end{scope}
\end{tikzpicture}
  \end{maybepreview}
 \caption{Our dual-camera setup on the UR10 arm used for the Amazon Picking Challenge 2016.}
 \label{fig:endeffector}
\end{figure}

Attached to the arm is a custom-built endeffector (see 
\cref{fig:endeffector}). For reaching into the deep and narrow APC shelf 
bins, we use a linear actuator capable of 37\,cm extension. On the tip 
of the linear extension, we mounted a rotary joint to be able to carry 
out both front and top grasps.

For grasping the items, we decided to employ a suction gripper. This 
choice was motivated by the large success of suction methods during the 
APC~2015~\citep{correll2016lessons}, and also due to the presented set 
of objects for the APC 2016, most of which could be manipulated easily 
using suction. Our suction system is designed to generate both high 
vacuum \textit{and} high air flow. The former is needed to lift heavy 
objects, the latter for objects on which the suction cup cannot make a 
perfect vacuum seal.

The suction cup has a 3\,cm diameter. For most objects, it is
sufficient to simply place the suction cup anywhere on the object.
For simple arrangements, this suction pose could be inferred from
a bounding box. For more complex arrangements with occlusions, there is an
increased risk of retrieving the wrong object. Similarly, very small objects
can be easily missed.
For these reasons, a high-quality localization such as offered by pixel-wise
semantic segmentation is required.
The final suction pose is derived using two different heuristics (top- or center grasp)
operating on the segmentation mask depending on the object height.
A nullspace-optimizing IK solver and keyframe interpolation are used to generate
motions, similar to the one described in \citet{schwarz2016drc}.
Further details on robot motion control and grasp generation are described
by \citet{schwarz2017apc}.

For control and computations, two computers are connected to the system. 
The first, tasked with high- and low-level control of the robot, is 
equipped with an Intel Core i7-4790K CPU (4\,GHz). The second is 
used for vision processing, and contains two Intel Xeon E5-2670 v2 
(2.5\,GHz) and four NVIDIA Titan X GPUs. For training, all four GPUs can 
be used to accelerate training time. At test time, two GPUs are used in 
parallel for the two deep learning approaches: object detection and
semantic segmentation (see \cref{sec:perception}).

\begin{figure*}[t]
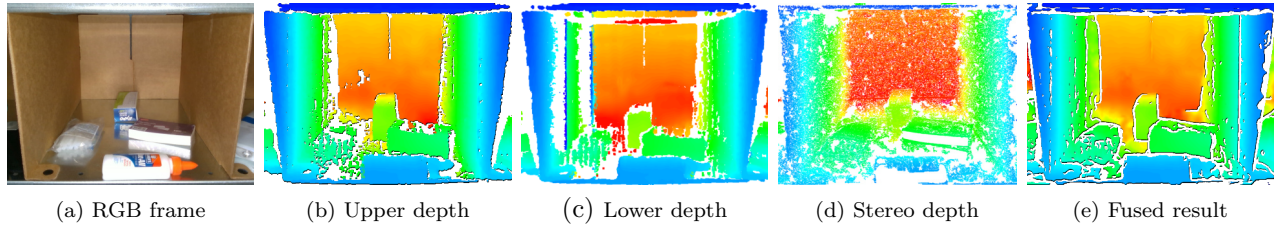

 \begin{maybepreview} \centering
 \setlength{\tabcolsep}{2pt}
 \begin{tabular}{ccccc}
  \includegraphics[height=.14\linewidth]{images/perception_examples/frame1/rgb_cropped} &
  \includegraphics[height=.14\linewidth]{images/perception_examples/frame1/cam2_filled_cropped_color} &
  \includegraphics[height=.14\linewidth]{images/perception_examples/frame1/cam1_filled_cropped_color} &
  \includegraphics[height=.14\linewidth]{images/perception_examples/frame1/stereo_cropped_color} &
  \includegraphics[height=.14\linewidth]{images/perception_examples/frame1/depth_filled_cropped_color} \\
  \footnotesize(a) RGB frame & \footnotesize(b) Upper depth & (c) \footnotesize Lower depth & \footnotesize (d) Stereo depth & \footnotesize (e) Fused result \\
 \end{tabular}
 \end{maybepreview}
 \caption{RGB-D fusion from two sensors.
 Note the corruption in the left wall in the lower depth frame, which is
 corrected in the fused result.
 }
 \label{fig:fusion}
\end{figure*}

\subsection{RGB-D Preprocessing}
\label{sec:rgbd-preprocessing}

After experimenting with multiple sensors setups in the APC setting, we 
decided to use the Intel RealSense SR300 RGB-D sensor due to its 
low weight, high resolution, and short-range capabilities. However, 
we noticed that the depth sensor produced systematic artifacts on the 
walls of the shelf (cf. \Cref{fig:fusion}(c)). The artifacts seem to depend on the viewing angle, 
i.e. they were present only on the right side of the image. To rectify 
this situation, we designed a dual sensor setup, with one of the sensors 
rotated $180^\circ$ (see \cref{fig:endeffector}). To reduce the undesirable
effect of external illumination, we used local LED lighting while capturing
the data.

Using two separate sensors also makes a second RGB stream available. To 
exploit this, we also compute dense stereo disparity between the two 
RGB cameras using LIB\-ELAS~\citep{Geiger2010ACCV}. The three depth 
streams are projected into one common frame (the upper camera in our 
case) and are then fused using a linear combination with predefined 
weights $\alpha_i$\footnote{Our experiments use $\alpha_{\text{stereo}}=0.1$ and $\alpha_{\text{RGB-D}}=40.0$.}.
In particular, 
the stereo stream is fused using a low weight, since the depth 
measurements of the SR300 cameras are usually more precise (but not 
always available). Finally, we distrust measurements where the different 
depth sources disagree by introducing an additional ``spread''
weight $w$.
In summary, we obtain the following equations for combining depth 
measurements $D_i$ of a single pixel to a depth measurement $D$ and weight $w$:
\begin{align}
 D &= \frac{\sum \alpha_i D_i}{\sum \alpha_i}, \\
 w &= \exp(-(\max_{i} D_i - \min_{i} D_i)).
\end{align}
Pixels where $\max_{i} D_i - \min_{i} D_i > 5\,\textrm{cm}$ are 
disregarded entirely. \Cref{fig:fusion} shows an exemplary scene with 
individual raw sensor measurements as well as the fused depth map.

Since the resulting fused depth map is usually sparse, we need to fill 
in the missing data. We follow the work of \citet{ferstl2013image}, who 
upsample depth images guided by a high-resolution grayscale image. In 
contrast to many other guided upsampling approaches, this one does not 
assume a regular upsampling grid. Instead, any binary (or even 
real-valued) weight matrix can be used to specify the location of source 
pixels in the output domain. This makes the approach applicable to our 
scenario, where the mask of valid pixels has no inherent structure.

The upsampling is formulated as an optimization problem, minimizing the energy
term
\newcommand*\diff{\mathop{}\!\mathrm{d}}
\begin{multline}
 \min_{u,v} \left\{ \alpha_1 \int_{\Omega_H} | T^{\frac{1}{2}} (\Delta u - v)| \diff x + \right. \\
   \left. \alpha_0 \int_{\Omega_H} | \Delta v | \diff x + \int_{\Omega_H} w |(u - D_s)|^2 \diff x \right\},
\end{multline}
where the first two summands regularize the solution using Total Generalized
Variation, and the last summand is the data term.
For details about the problem formulation and the solver algorithm, we refer
the reader to \citet{ferstl2013image}.
The guided upsampling was implemented in CUDA to achieve near real-time 
performance ($<$100\,ms per image).

\subsection{Overall Results}

The system performed both the picking and stowing 
task successfully during the APC 2016. 

\subsubsection{Stowing task.}
Our system was able to stow
eleven out of twelve items into the shelf.\footnote{Video at 
\url{https://youtu.be/B6ny9ONfdx4}} 
However, one of the successfully 
stowed items was misrecognized: Our approach falsely identified a whiteboard eraser as a toothbrush
and placed it into the shelf.
This meant that the system could not 
recognize the toothbrush as the final item remaining in the tote.
This unlikely event was expected to be caught by a built-in fallback
mechanism which would attempt to recognize all known 
objects. However, this mechanism failed because the only remaining
object was thin and therefore discarded based on a size threshold. The
misrecognition of the item led to the attainment of the second place in
the stow task.

\begin{table}
 \centering
 \begin{threeparttable}
  \caption{Picking Run at APC 2016}
  \begin{tabular}{cp{2.4cm}ccc}
    \toprule
    Bin & Item & Pick & Drop & Report \\
    \midrule
    A & duct tape    & $\times$     & $\times$      & $\times$ \\
    B & bunny book   & $\checkmark$ & $\checkmark$  & $\times$\tnote{2} \\
    C & squeaky eggs & $\checkmark$ & $\times$      & $\checkmark$ \\
    D & crayons\tnote{1} & $\checkmark$ & $\times$      & $\checkmark$ \\
    E & coffee       & $\checkmark$ & $\checkmark$  & $\times$\tnote{2} \\
    F & hooks        & $\checkmark$ & $\times$      & $\checkmark$ \\
    G & scissors     & $\times$     & $\times$      & $\times$ \\
    H & plush bear   & $\checkmark$ & $\times$      & $\checkmark$ \\
    I & curtain      & $\checkmark$ & $\times$      & $\checkmark$ \\
    J & tissue box   & $\checkmark$ & $\times$      & $\checkmark$ \\
    K & sippy cup    & $\checkmark$ & $\times$      & $\checkmark$ \\
    L & pencil cup   & $\checkmark$ & $\checkmark$  & $\times$\tnote{2} \\
    \midrule
      & Sum          & 10           & 3             & 7 \\
    \bottomrule
  \end{tabular}
  \begin{tablenotes}
   \item [1] Misrecognized, corrected on second attempt.
   \item [2] Incorrect report, resulting in penalty.
  \end{tablenotes}
 \end{threeparttable}
\end{table}

\subsubsection{Picking task.}
In the picking task, our system picked ten out of twelve 
items.\footnote{Video at \url{https://youtu.be/q9YiD80vwDc}} Despite the
high success rate (the winning Team DELFT achieved a success pick-up 
rate of only nine items),  only a third place was achieved as a 
consequence of  dropping three items during picking. 
Note that we could correctly
recognize that the objects were not picked successfully using the air velocity sensor of our robot.
However, the system incorrectly deduced 
that the items were still in the shelf, when they actually dropped over 
the ledge and into the tote. Since the system was required to deliver a 
report on the final object locations, the resulting penalties reduced 
our score from 152 points to 97 points---just behind the first and 
second place of Team DELFT and PFN, both of which achieved 105 points.

\subsection{Object Detection and Semantic Segmentation}
\label{sec:eval:object_detection}

\begin{figure}
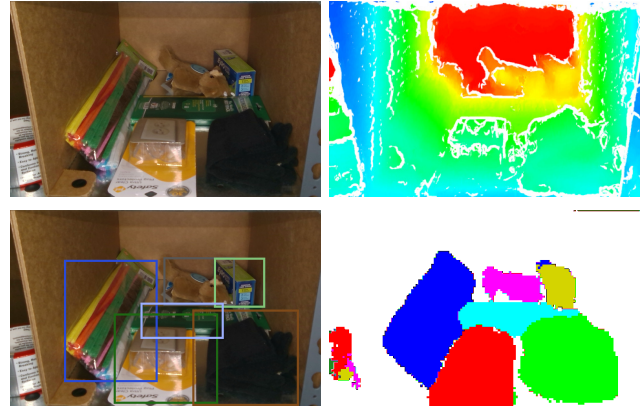

 \begin{maybepreview} \centering

 \includegraphics[width=.49\linewidth]{images/perception_examples/frame0/rgb_crop_resized}
 \includegraphics[width=.49\linewidth]{images/perception_examples/frame0/depth_colorized_crop}
 
 \vspace{1ex}
 
 \includegraphics[width=.49\linewidth]{images/perception_examples/frame0/densecap_lua}
 \includegraphics[width=.49\linewidth]{images/perception_examples/frame0/segmentation_crop}
 \end{maybepreview}

 \caption{Object perception example. Upper row: Input RGB and depth frames.
 Lower row: Object detection and semantic segmentation results (colors are not
 correlated).}
 \label{fig:perception:example}
\end{figure}

In addition to the system-level evaluation at the APC, we evaluated our 
perception approaches on our own annotated dataset, which was also used 
for training the final APC model. The dataset contains 201 shelf frames, and 132 
tote frames. The frames vary in the number of objects and location in 
the shelf. As far as we are aware, this number of frames is quite low in 
comparison to other teams~\citep[e.g. Team Princeton-MIT,][]{Zeng:2017:ICRA},
 which highlights the effectiveness of our 
transfer learning approach. \Cref{fig:perception:example} shows an 
exemplary scene from the dataset with object detection and segmentation 
results.

For evaluation, we define a five-fold cross validation split on the 
shelf and tote datasets.
To ensure that the examples for each
class are distributed as evenly across the splits as possible, we use
Iterative Stratified Sampling~\citep{sechidis2011stratification}.

Please note that the DenseCap pipeline has been thoroughly evaluated on the
Visual Genome Dataset by \citet{Johnson:2016:CVPR} and the semantic segmentation
network has been evaluated on the NYU Depth v2 dataset by \citet{Husain:2016:RAL}.
Both achieved state-of-the-art results on the respective datasets.

\subsubsection{Object detection.}

\begin{table*}[!t]
 \centering
 \pgfplotstableread{results/apc_results.txt}\apcresults
\begin{threeparttable}
 \caption[Object detection architectures]{   Evaluation of object detection architectures on the shelf dataset.
   The mAP object detection score and the F1 localization score are shown for each architecture.
 }
 \label{tab:apc:architectures}
 \pgfplotstabletypeset[
  highlight col max={\apcresults}{informed_mAP},
  highlight col max={\apcresults}{uninformed_mAP},
  highlight col max={\apcresults}{F1},
  columns/{Dataset}/.style={string type,
    column name={Input},
    column type=l,
    string replace={depth-hhav}{RGB-D (TGV)},
    string replace={depth-upper-hhav}{RGB-D (single)\tnote{1}},
    string replace={merged}{RGB},
    string replace={depth-old}{RGB-D (DT)\tnote{2}},
  },
  columns/{Variant}/.style={string type,
    column type=l,
    string replace={notailor}{SVM (plain)},
    string replace={tailor}{SVM (tailor)},
    string replace={hha_concat}{HHA Features (Fig.~\ref{fig:method:depth:downsample})},     string replace={hha_cnn}{HHA CNN (Fig.~\ref{fig:method:depth:2cnn})},     string replace={normal}{Softmax (with augmentation)},
    string replace={noaug}{Softmax (no augmentation)},
    string replace={cross}{Distillation},     string replace={proposal}{Ext. Proposals},   },
  columns/{uninformed_mAP}/.style={
    column name={Uninf.},
    zero to dash,
    fixed zerofill,
    multiply with=0.01,
    precision=3,
  },
  columns/{informed_mAP}/.style={
    column name={Inf.},
    fixed zerofill,
    multiply with=0.01,
    precision=3,
  },
  columns/{F1}/.style={
    fixed zerofill,
    multiply with=0.01,
    precision=3,
  },
  every head row/.style={
    before row={
      \toprule
       & & \multicolumn{2}{c}{mAP} \\
       \cmidrule(r){3-4}
    },
    after row=\midrule
  },
  every last row/.style={after row=\bottomrule},
  every row no 1/.style={after row=\midrule},
  every row no 3/.style={after row=\midrule},
  every row no 7/.style={after row=\midrule},
 ]{\apcresults}
 \begin{tablenotes}
  \item[1] Without depth fusion, only from upper camera. Filled using TGV method.
  \item[2] Old filling method used during APC 2016 based on a color-guided smoothing filter.
 \end{tablenotes}
 \end{threeparttable}
  \end{table*}

Traditionally, object detectors are evaluated using a retrieval metric like
\textit{mean Average Precision (mAP)}, which measures the quality of ranked
results over the whole test set. Usually, an Intersection-over-Union (IoU) threshold of
0.5 is chosen, focusing on detection and rough localization instead of precise
localization.
Generally, the mAP metric as defined above places greater weight on correct
detection than on precise localization. Indeed, it is much easier to achieve
perfect average precision scores than perfect localization precision.

To also provide location sensitivity, one can define a metric for
object detection based on pixel-level precision and recall. In this work,
we consider the use case for an object detector in the context of
warehouse automation: It is known that a particular object resides in a particular
shelf bin, and we need to retrieve it. Here, we are only interested in the
detection $i$ with maximum confidence $c_i$ for this object class. We measure
its precision and recall as follows:
\begin{align}
 \textrm{precision} &= \frac{|B \cap G|}{|B \cup G|} = \textrm{IoU}(B,G) \\
 \textrm{recall}    &= \frac{|B \cap G|}{|G|},
\end{align}
where $B$ is the detected bounding box and $G$ denotes the closest ground truth
bounding box. Note that a complete mislocalization results in zero precision
and recall.
Both quantities are then combined into the final F1 score:
\begin{align}
 \textrm{F1} &= 2 \cdot \frac{\textrm{precision} \cdot \textrm{recall} }{\textrm{precision} + \textrm{recall} }.
\end{align}

\Cref{tab:apc:architectures} shows the influence of the design choices described
in \cref{sec:object-detection}. Both mAP (informed and uninformed case) and the custom
F1 metric are shown for each architecture. As a first result we note that the softmax variant
with its ability to finetune the entire network including region proposal is
far superior to a fixed network paired with an SVM classifier. In particular,
the SVM classifier is bad at ranking the detections across images, which is
evident in the mAP metric. A calibration step (e.g. Platt scaling) could
improve this behavior.
Note that the F1 score, which is more relevant in the APC scenario, is closer
to the rest of the methods, but still suboptimal.
Training the SVM on-the-fly for just the items in the current shelf bin
(``tailor'' variant) makes little difference in both metrics.
All remaining tests were performed with the superior softmax classifier.
Data augmentation (image mirroring) slightly improves performance,
so all other tests were performed with augmentation.

As expected, incorporating depth measurements results in increased performance.
The external RGB-D proposal generator performs better than the naive HHA
concatenation.
However, reusing the RGB CNN for depth feature computation outperforms the
proposal generator.
Finally, training a depth CNN using Cross Modal Distillation gives the best
results.

\begin{table*}[ht]
 \centering
 \begin{threeparttable}
  \caption{Perception runtimes.}
  \label{tab:timings}
  \begin{tabular}{lrrrrr}
    \toprule
    Phase &  \multicolumn{4}{c}{Object detection}                                & Segmentation \\
    \cmidrule(lr){2-5}
          & RGB-D proposal & SVM               & Softmax (RGB) & Softmax (RGB-D) & \\
    \midrule
    Train &        -       &        -          &    45\,min    & 4.5\,h          & $\approx$ 5\,h \\
    Test  & 1006\,ms       & 3342\,ms\tnote{1} &   340\,ms     & 400\,ms         & $\approx$ 900\,ms \\
    \bottomrule
  \end{tabular}
  \begin{tablenotes}
   \item [1] Includes just-in-time SVM training
  \end{tablenotes}
 \end{threeparttable}
\end{table*}

We also investigate the depth fusion method described in
\cref{sec:rgbd-preprocessing}. The TGV-regularized method is superior to
a simple color-guided smoothing filter. The advantage of fusing the two camera
streams can be seen.
\begin{table}
 \centering
 \caption[Final results]{Final object detection results on the APC dataset.}
 \label{tab:detection:final}
 \begin{tabular}{lrrr}
  \toprule
                     & \multicolumn{2}{c}{mAP} & F1 \\
  \cmidrule(lr){2-3}
  Dataset            & Uninformed & Informed   & \\
  \midrule
  Shelf              & 0.878      & 0.912      & 0.798 \\
  Tote               & 0.870      & 0.887      & 0.779 \\
  \bottomrule
 \end{tabular}
\end{table}
\pgfplotstableread{results/apc_breakdown.txt}\apcbreakdown
\pgfplotstablesort[sort key=F1]\apcbreakdownf{\apcbreakdown}
\begin{figure}
 \begin{maybepreview} \begin{tikzpicture}
  \begin{axis}[ybar,
    width=\linewidth,
    height=4cm,
    ymin=0.4,
    ymax=1.2,
    xmin=-1,
    xmax=39,
    xtick=\empty,
    bar width=1,
    line width=0,
    ylabel={F1 Score},
    ytick pos=left,
    title={Object Detection},
    title style={yshift={-1ex}}
    ]
    
    \addplot+[bar width=1] table[y=F1,x expr=\coordindex] {\apcbreakdownf};
    \node[draw=red,line width=0.5pt,circle,text width=0.65cm,pin={[align=left,font=\footnotesize,pin distance=1ex]+75:{\texttt{red toothbrush}\\\texttt{glue bottle}}}] at (axis cs:0.5, 0.6) {};
    \node[draw=red,line width=0.5pt,circle,text width=0.2cm,pin={[align=right,font=\footnotesize]+165:{\texttt{tissue box}}}] at (axis cs:38, 0.94) {};
  \end{axis}
 \end{tikzpicture}
 \end{maybepreview}
 \begin{tikzpicture}
\begin{axis}[ybar,
	width=\linewidth,
    height=4cm,
    ymin=0.4,
    ymax=1.2,
    xmin=-1,
    xmax=39,
    xtick=\empty,
    bar width=1,
    line width=0,
    ylabel={F1 Score},
    ytick pos=left,
title={Semantic Segmentation},
title style={yshift={-1ex}}
    ]
\addplot table[header=false,x expr=\coordindex,y index=1,col sep=comma] {
fiskars scissors red, 0.4780                
oral b toothbrush red, 0.6790 
dove beauty bar, 0.7400                       
oral b toothbrush green, 0.7460 
creativity chenille stems , 0.7610             
fitness gear 3lb dumbbell, 0.7630         
up glucose bottle, 0.7860 
cool shot glue sticks, 0.7930                 
dasani water bottle, 0.7960                  
scotch duct tape, 0.8100 
crayola 24 ct, 0.8110                         
cherokee easy tee-shirt, 0.8210
clorox utility brush, 0.8210
expo dry erase board eraser , 0.8240          
safety first outlet plugs, 0.8340 
cloud b plush bear, 0.8360
elmers washable no run school glue , 0.8380    
soft white lightbulb, 0.8410 
staples index cards, 0.8470 
rawlings baseball, 0.8580 
kyjen squeakin eggs plush puppies, 0.8600 
command hooks, 0.8770 
barkely hide bones, 0.8780
easter turtle sippy cup , 0.8790             
womens knit gloves, 0.8870 
kleenex paper towels, 0.8880
woods extension cord, 0.8880
hanes tube socks, 0.8890                       
jane eyre dvd, 0.8930                    
laugh out loud joke book, 0.8940 
scotch bubble mailer, 0.8940 
folgers classic roast coffee, 0.8980 
i am a bunny book, 0.9060       
peva shower curtain liner, 0.9160 
platinum pets dog bowl, 0.9170 
rolodex jumbo pencil cup, 0.9260 
dr browns bottle brush, 0.9400               
kleenex tissue box, 0.9420 
ticonderoga 12 pencils, 0.9500
};

\node[draw=red,line width=0.5pt,circle,text width=0.2cm,pin={[align=left,font=\footnotesize,pin distance=1cm]+85:{\texttt{fiskars scissors}}}] at (axis cs:0, 0.48) {};
\node[draw=red,line width=0.5pt,circle,text width=0.2cm,pin={[align=right,font=\footnotesize]+165:{\texttt{pencils}}}] at (axis cs:38, 0.95) {};

\end{axis}
\end{tikzpicture}  \caption{F1 score distribution over the objects for object detection (top) and semantic segmentation (bottom).
 Results are averaged over the cross validation splits.
 For object detection, the best RGB-D model is used.
 }
 \label{fig:densecap:distribution}
\end{figure}
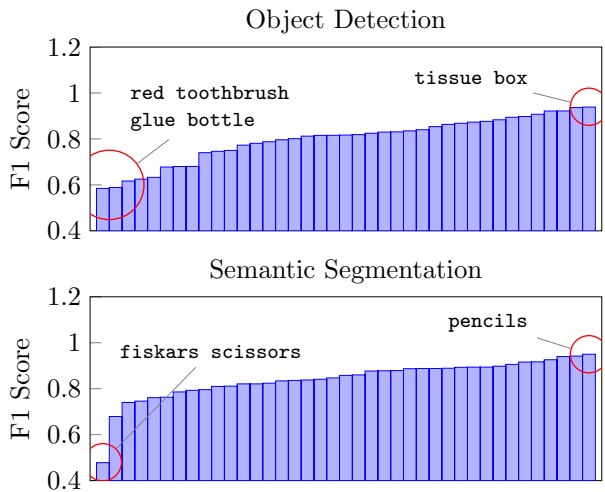
\Cref{tab:detection:final} shows the final object detection results on the
shelf and tote datasets.

\begin{table}[th]
 \centering
 \begin{threeparttable}
  \caption{F1 scores for semantic segmentation.}
  \label{tab:perception-segmentation:scores}
 \begin{tabular}{lrrrr}
  \toprule
           & \multicolumn{2}{c}{Shelf} & \multicolumn{2}{c}{Tote} \\
  \cmidrule(lr){2-3} \cmidrule(lr){4-5}
  Method                     & Uninf. & Inf.     & Uninf.     & Inf. \\
  \midrule
  Raw depth                  & 0.713  & 0.735    & -          & - \\
  HHA depth                  & 0.780  & 0.813    & 0.817      & 0.839 \\
  \midrule
  Det+Seg\tnote{1}       & 0.795  & 0.827    & 0.831      & 0.853 \\
  \bottomrule
 \end{tabular}
 \begin{tablenotes}
 \item [1] Object Detection + Segmentation.
 \end{tablenotes}
 \end{threeparttable}
\end{table}

\subsubsection{Semantic Segmentation.}

For segmentation,
pixel-level precision and recall are calculated. Resulting F1 scores are 
shown in \cref{tab:perception-segmentation:scores}.
Knowledge of the set 
of possible objects improves the performance slightly but consistently.
The chosen HHA encoding of depth is far superior to raw depth
($\sim 7\%$ increase).
Finally, the combination of the finetuned object detector and 
the semantic segmentation yields a small but consistent increase in 
performance.

\Cref{fig:densecap:distribution} shows the distribution 
of difficulty across all object classes. 
Both methods struggle mostly with small and/or shiny
objects (tooth brush, scissors), the semantic segmentation even more so,
reaching an F1 score below $0.5$. 
We believe that both
methods are affected by insufficient image resolution and annotation errors,
which result in a greater effect on objects of smaller size.
The object detection also struggles with elongated shapes (e.g. toothbrush),
where the approximation of the contour as a bounding box may be deficient
under certain viewing angles.

We also measured the runtime of the
different modules in our setup (see \cref{tab:timings}).
The training time for the RGB-D object detection pipeline is much longer,
due to high memory utilization on the GPU. This could probably be improved
e.g. through precomputation of the initial pretrained layers, however training
time was not taken into consideration in our application.
At test time, all approaches achieve sufficient runtimes ($\leq 1$\,s)
for bin-picking applications.
Note that object detection and semantic segmentation usually run in parallel.

\section{Application to Disaster Response}
\label{sec:centauro}

\begin{figure*}
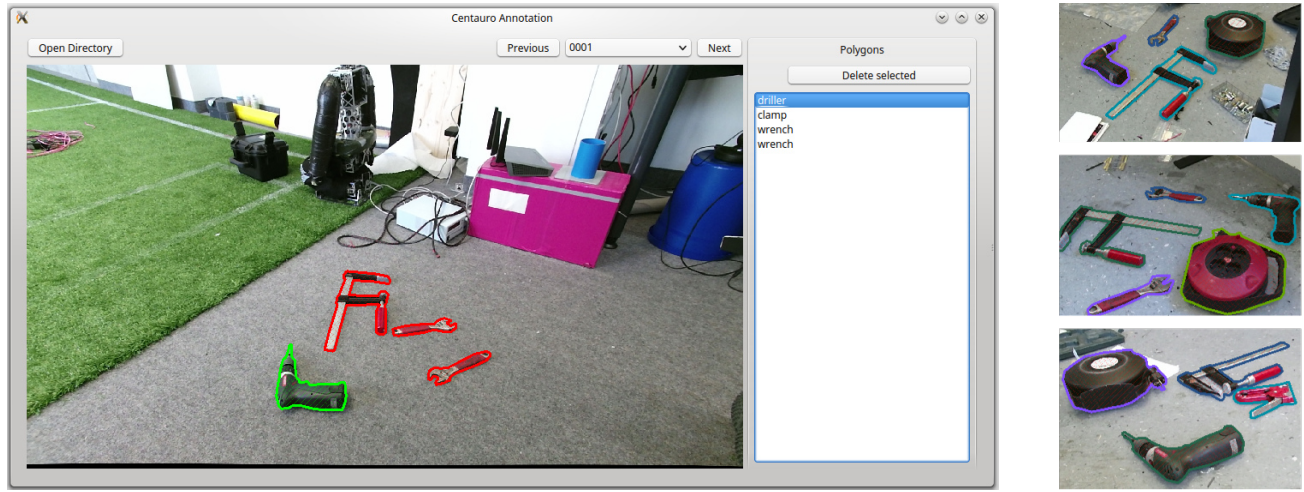

 \begin{maybepreview}	\centering
	\includegraphics[height=0.375\linewidth] {images/centauro_dataset/centauro_annotation}
	\hfill
	\includegraphics[height=0.375\linewidth] {images/centauro_dataset/anno_crops}
 \end{maybepreview}
	\caption{Our annotation tool (left) and three exemplar cropped frames from the captured dataset (right).}
	\label{fig:annotool}
\end{figure*}

 \begin{table*}[th]
 \centering
 \caption{Object detection results on the \acro dataset.}
 \label{tab:densecap:tools}
 \small
 \begin{tabular}{r@{$\times$}lc@{/}cc@{/}cc@{/}cc@{/}cc@{/}cc@{/}cc@{/}c}
  \toprule
  \multicolumn{2}{c}{Resolution} & \multicolumn{2}{c}{Clamp} & \multicolumn{2}{c}{Door handle} & \multicolumn{2}{c}{Driller} & \multicolumn{2}{c}{Extension} & \multicolumn{2}{c}{Stapler} & \multicolumn{2}{c}{Wrench} & \multicolumn{2}{c}{Mean}\\
  \cmidrule(lr){3-4} \cmidrule(lr){5-6} \cmidrule(lr){7-8} \cmidrule(lr){9-10} \cmidrule(lr){11-12} \cmidrule(lr){13-14} \cmidrule(lr){15-16}
  \multicolumn{2}{c}{}           & AP & F1                   & AP & F1                         & AP & F1                     & AP & F1                       & AP & F1                     & AP & F1                    & AP & F1 \\
  \midrule
    720 & 507 & 0.881 & 0.783 & 0.522 & 0.554 & 0.986 & 0.875 & 1.000 & 0.938 & 0.960 & 0.814 & 0.656 & 0.661 & 0.834 & 0.771 \\
    1080 & 760 & 0.926 & 0.829 & 0.867 & 0.632 & 0.972 & 0.893 & 1.000 & 0.950 & 0.992 & 0.892 & 0.927 & 0.848 & 0.947 & 0.841 \\
    1470 & 1035 & 0.913 & 0.814 & 0.974 & 0.745 & 1.000 & 0.915 & 1.000 & 0.952 & 0.999 & 0.909 & 0.949 & 0.860 & 0.973 & 0.866 \\

  \bottomrule
 \end{tabular}
 
\end{table*}

\begin{table*}[th]
	\centering
	\caption{Class-wise pixel classification F1 score of the semantic segmentation method in disaster-response environments.}
	\begin{tabular}{ccccccc|r}
	  \toprule
	  Clamp  & Door handle & Driller & Extension & Stapler & Wrench & Background & Mean\\
	  \midrule
	  0.727 & 0.751 & 0.769 & 0.889 & 0.775 & 0.734 & 0.992 & 0.805 \\

	  \bottomrule
	\end{tabular}
	\label{segmentation_table}
\end{table*}

To show the general applicability of the developed pipeline in other contexts,
we also evaluate it in a second, unrelated domain. Note that the pipeline
was initially designed for bin-picking perception, and is now merely adapted to this
new scenario.

The dataset that was used for validating our approach was captured in the European
\acro\footnote{\url{https://www.centauro-project.eu}} project, which aims
to develop a human-robot symbiotic system for disaster response.
To reduce the work load of the human operator, several perception and
manipulation capabilities should be done autonomously by the robot
system. For example, the robot should be able to identify commonly used tools
 like wrenches or drillers, correctly grasp and utilize them. Here, we
 apply the techniques that were used for bin-picking during the APC to
 the disaster response scenario.
The dataset consists of
127 manually annotated \mbox{RGB-D} frames, captured in a cluttered mechanics 
workshop with six different object classes: Five mechanic tools (clamp, driller,
extension box, stapler, wrench) and door handles.
The dataset was captured with a Kinect version 2 camera and annotated manually 
with a tool developed in-house. Figure \ref{fig:annotool} shows a screenshot of the 
annotation tool and examples of three annotated frames.
In addition to the unique setting, the dataset differs from common RGB-D datasets
by offering pixel-wise labeling, highly cluttered background, and the high
capture resolution of the Kinect v2 camera.

As in \cref{sec:eval:object_detection}, we define a five-fold cross validation
using Iterative Stratified Sampling~\citep{sechidis2011stratification} for
evaluating object detection and semantic segmentation. Since Kinect v2
does not measure depth in the margin areas (especially left and right borders)
of the $1920\times1080$ RGB image, we crop the frames to the area where depth
measurements are available ($1470\times1035$).

\subsection{Detection and Segmentation}
\label{sec:centauro-segmentation}
The object detection pipeline is identical to the one described in \cref{sec:object-detection}.
However, since there are frames in the dataset that are either very cluttered,
making it harder to robustly estimate a ground plane, or have no ground plane
in the image at all (see \cref{fig:seg_result}, bottom), we do not use the geometric
HHA features here. Instead, we train the depth CNN using Cross Modal Distillation
on raw depth.
The segmentation network is also very similar to the one illustrated in \cref{fig:segmentation-cnn}.
Here, we feed raw depth (replicated to the three input channels) instead of HHA into the depth branch.

\subsection{Evaluation}
\label{sec:centauro-evaluation}

\begin{figure*}
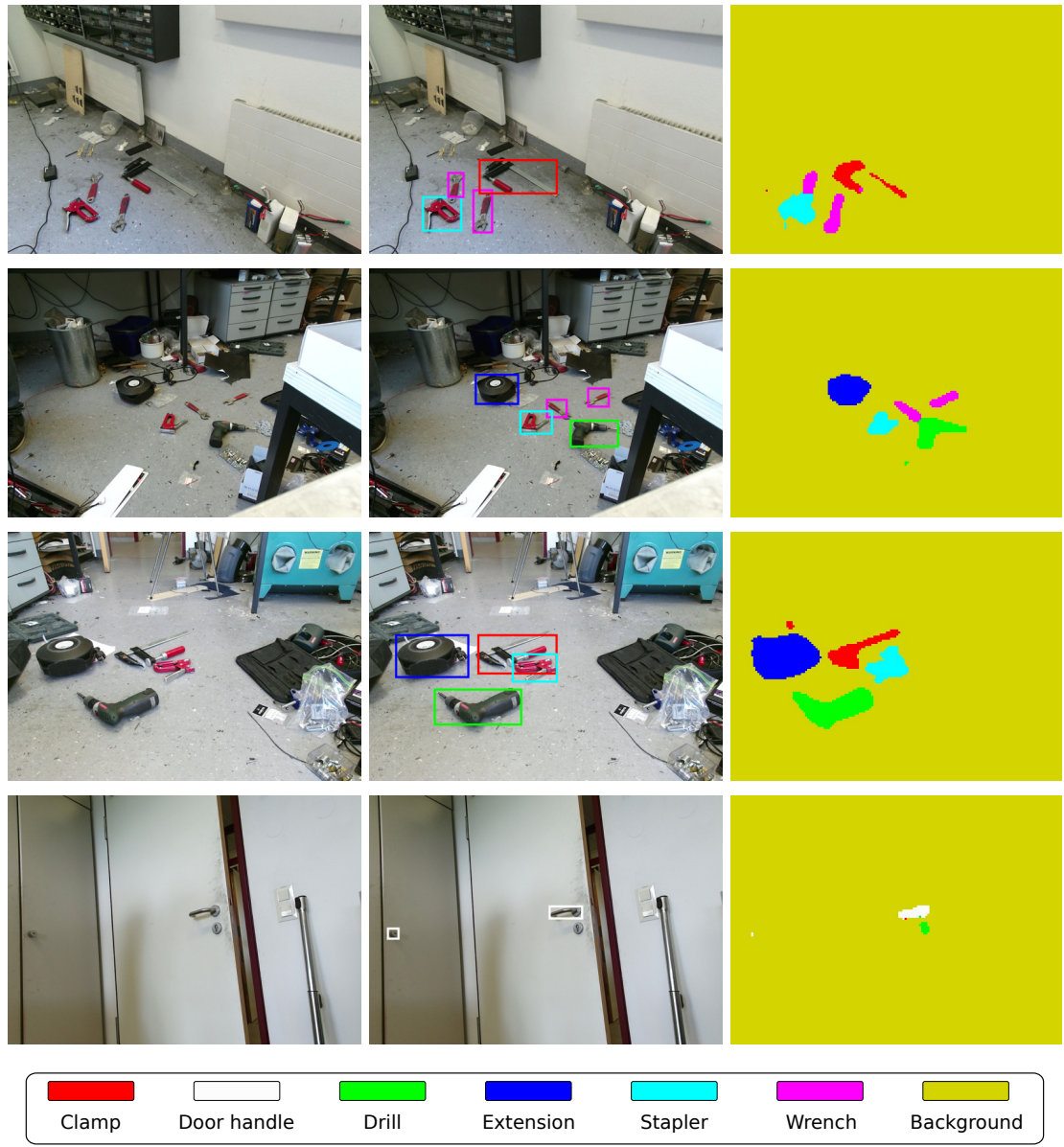

 \centering
 \begin{maybepreview} \centering
 \includegraphics[height=.2\linewidth]{images/centauro_examples/img002/input}
 \includegraphics[height=.2\linewidth]{images/centauro_examples/img002/out4}
 \includegraphics[height=.2\linewidth]{images/centauro_examples/img002/CR_0_03_0006}\\[.5em]
 \includegraphics[height=.2\linewidth]{images/centauro_examples/img004/input}
 \includegraphics[height=.2\linewidth]{images/centauro_examples/img004/out5}
 \includegraphics[height=.2\linewidth]{images/centauro_examples/img004/CR_0_02_0005}\\[.5em]
 \includegraphics[height=.2\linewidth]{images/centauro_examples/img010/input}
 \includegraphics[height=.2\linewidth]{images/centauro_examples/img010/out4}
 \includegraphics[height=.2\linewidth]{images/centauro_examples/img010/CR_0_06_0002}\\[.5em]
 \includegraphics[height=.2\linewidth]{images/centauro_examples/img024/input}
 \includegraphics[height=.2\linewidth]{images/centauro_examples/img024/out2}
 \includegraphics[height=.2\linewidth]{images/centauro_examples/img024/CR_0_07_0006}

 \begin{center}
 \includegraphics[width=.82\linewidth]{images/centauro_examples/seg_legend}
 \end{center}
 \end{maybepreview}
 \caption{Example scenes (left) and results of object detection (center) and semantic segmentation (right)
  on the disaster response dataset. For object detection, the highest resolution model is used.}
 \label{fig:seg_result}
\end{figure*}

Object detection results for the \acro dataset are shown in \cref{tab:densecap:tools}.
With the configuration from the APC dataset, the detector shows an acceptable mAP of
83.4\%.
However, by default the DenseCap pipeline scales the input images such that the largest
side is 720 pixels---reducing the available resolution. Since there are very small
objects that occupy as little as $4\%$ of the image width (e.g. door knobs, see \cref{fig:seg_result}, bottom row),
higher resolutions increase the performance by a large margin
(see \cref{tab:densecap:tools}), of course at the cost of training and prediction time.
Increasing the input size to the highest available resolution (1470$\times$1035)
yields near-perfect detection performance ($97.3\%$ mAP).
This increase is equally visible in the localization score, suggesting
that mis-localization (and not mis-detection) is the main cause for the low mAP at
lower resolution.
The full resolution model takes about 10\,h to train and has longer prediction times
of around 1\,s per image.
An intermediate resolution yields $94.7\%$ mAP with a prediction time of 550\,ms.

The segmentation results are presented in \cref{segmentation_table}.
Here the segmentation achieves good results, which is notable in the
presence of highly cluttered background with many other objects that
are visually rather similar. Finally, 
\Cref{fig:seg_result} shows an example result of our detection and segmentation approaches
on the \acro dataset.
Surprisingly, the highly cluttered background does not affect
the overall performance. Rather, the main difficulty with this dataset is the small
object size w.r.t. to the image size, in contrast to the APC data.

\section{Lessons Learned}
\label{sec:lessons-learned}

Designing the system, participating in the APC 2016, and finally the experiments
on disaster scenes was a valuable learning experience for us. We summarize and
discuss some of the points related to the perception system here.

First of all, our transfer learning approach has shown to work effectively in real-life
robotic applications, requiring only few
annotated images. We expect that further work (which is necessary for the ARC
2017) will reduce the number of required images even further, paving the way for
one-shot or few-shot learning.
The main insight here is that these problems can be approached successfully
with deep learning techniques, even if the amount of training data is low.
Furthermore, our experiments showed that it is beneficial to finetune pretrained architectures rather than relying on classical CNN+SVM combination in the considered scenarios, despite the limited training data.

One valuable lesson from the disaster response scenario is that input resolution
is an important parameter for the performance of detection approaches. Sadly,
this is currently a highly domain-specific parameter, requiring careful adaptation
to the task at hand---and the processing time available.
More generally, state-of-the-art deep learning techniques still require
significant amounts of manual tuning to adapt the models to the target domain.
Despite the replacement of handcrafted features with learned ones,
hyperparameters have to be chosen carefully to guarantee success.
Finally, tricks such as the HHA encoding may boost the performance in certain settings, 
but are not generally applicable to all domains.

\section{Conclusion}
\label{sec:conclusion}

We presented a successful adaptation of two different deep-learning-based
image understanding methods to robotic perception. Our experiments showed that by exploiting
transfer learning, such approaches can be applied to real-world manipulation
tasks without excessive need for annotating training images. 
We demonstrated their performance in two different settings. One is
a bin-picking scenario, carried out in the context of the Amazon Picking Challenge
(APC) 2016, where the number of categories, narrow working spaces, as well as
shiny and textureless surfaces pose major challenges. The APC 2016 was our
very first attempt to apply deep-learning techniques in a live robotic system.
Our team's success underlines the effectiveness of such methods in practice.
We believe that the reduction of training data is a key factor in such scenarios.
This will become even more crucial in future editions of the APC, where the
number of categories will increase and some categories may not even be known
during the training phase.

The second application scenario is disaster response. Here, the
main challenges for perception include severe clutter and 
unstructured background. Nevertheless, we showed that similar ideas
can be transferred to this setting. To validate this claim, we collected and
annotated a domain-specific dataset and observed encouraging performance
in both detection and segmentation tasks.

In future, we plan to bring both tasks closer together by integrating them
into a single network architecture. This would allow for end-to-end training
of both components simultaneously. Finally, we make the entire code
base used for the APC as well as the collected and annotated data
publicly available\footnote{\url{http://ais.uni-bonn.de/apc2016/}}.
We hope this will encourage other researchers to apply
deep models in live robotic systems.

\bibliographystyle{SageH}
\bibliography{references,long,refs-anton,ref-new}

\end{document}